\documentclass[a4paper]{paper}
\usepackage[landscape]{geometry}

\usepackage{graphicx}
\usepackage{float}
\usepackage{subfig}

\begin{document}

%
\title{Mapping Learning Algorithms on Data, a useful step for optimizing performances and their comparison}

\author{Filippo Neri, DIETI, University of Naples, Italy. email: filippo.neri.email@gmail.com}

 \maketitle

\begin{abstract}
In the paper, we propose a novel methodology to map learning algorithms on data (performance map) in order to gain more insights in the distribution of their performances across their parameter space. This methodology provides useful information when selecting a learner's best configuration for the data at hand, and it also enhances the comparison of learners across learning contexts.

In order to explain the proposed methodology, the study introduces the notions of learning context, performance map, and high performance function.
It then applies these concepts to a variety of learning contexts to show how their use
can provide more insights in a learner's behavior, and can enhance the  comparison of learners across learning contexts.

The study is completed by an extensive experimental study describing how the proposed methodology can be applied.
\end{abstract}


\section{Introduction}
The standard approach used in machine learning is to compare learning algorithms consists in contrasting their performances on  a data set unseen during the learning phase. A learner's performance is expressed in the form of a single numeric value representing, for instance, its accuracy, the error rate, etc.. Usually a confidence interval around the mean performance value is also provided.
However, in the end, a whole learner behavior is condensed into just one single number (i.e. the mean accuracy). 

All other information about the learning process (i.e. how the search in the hypothesis space was conducted, what effect changing learning parameters produces, how human readable is the found concept, etc.)  is simply discarded.
From the theoretical point of view, the user is then supposed to select a learner over the other just by considering a single number.

On the opposite, from the practical point of view, the literature papers may only partially helpful as they usually hide away the important step of \emph{parameter selection} that is, however, performed by the authors but generally not discussed in the paper.

When considering real data, we believe, instead, that a) the step of parameter selection should be considered a full part of the learning process, and that b) a learner's parameter sensitivity should play
a role in comparing learners across different learning contexts.  
In fact, if a learner's result is very sensitive to its settings, the user may want to consider selecting a lower performing learner with stabler results to ensure a more robust behavior on future data.

Following the above considerations, this study describes a new methodology to compare learning systems by using \emph{performance maps} that makes explicit a learner's sensitivity to its parameter settings.

We define a \emph{performance map} as the set of performance values, associated to the parameter settings
that produced them, when a leaner is applied to some data. 
\emph{Performance maps} are functions of \emph{learning contexts}.
In order to understand how to build them, let us then define what a \emph{learning context} is for the extent of this study.

A \emph{learning context LC} is a quadruple made of: 
\begin{enumerate}
	\item a learning algorithm $L$,
	\item a meta optimization method $M$, 
	\item the meta-optimized parameter space $MOPS$: the set of parameter settings for $L$ considered during meta-optimization, and
	\item a data set D.
\end{enumerate}

Meta-optimization of learning systems, or hyper-parameter learning, or meta-learning consists 
in finding the best performing parameter settings for a learner by searching the space of all possible parameter settings
 \cite{Blum2003,grefenstette1986,Eiben1999,ReifMetaLearning,Feurer2019,Ribalta2017}. 

Meta-optimization of a learner $L$ is achieved by performing multiple runs of $L$ on $D$, using several parameter settings, in order to evaluate $L$'s performance for each considered parameter settings. 
Either exhaustive search or a specific meta-optimization algorithm $M$ can be used.
And, the set of $L$'s parameter settings evaluated during the meta-optimization process is
the \emph{meta-optimized parameter space} ($MOPS$).
The collection of pairs  \mbox{< s, $L$'s performance >}, with $s$ in $MOPS$, allows to create 
the \emph{performance map(LC)} that we are interested.
 
The selected meta-optimization method $M$  determines the composition of $MOPS$ and, in turns, of the \emph{performance map(LC)}. Performance maps can be either complete, if $MOPS$ is equal to the set of all parameter settings for $L$,  
or partial/approximated, if $MOPS$ is a proper subset of it.

Meta-optimization is very effective and can improve significantly the performance of a learning algorithm \cite{Camilleri2014203,Camilleri2014582}. 
We will show some instances of the case in the experimental part below.
In this paper, however, we do not focus on meta-optimization per se but we use it only as a tool to build performance maps.

In the description of how performance maps are created, the machine learner expert can easily
recognize a formalized version of the manual parameter tuning process accomplished by all authors in order to select the 'most suitable' configuration for running the learners discussed in their papers.

Novelties of this paper include:
\begin{enumerate}
\item 
the notion of \emph{learning context} and it use to compare learning algorithms or to tune their performance.
\item
the definition of \emph{performance maps} and how they can be used to compare learners
\item 
the description of how to create approximate (partial) performance maps with relatively low computational cost yet providing 'satisfactory' information
\item
the suggestion  that previous research in the literature, has been implicitly using a weak version of the performance maps method, here described, usually performed informally by the authors before selecting the configuration to use in the learners discussed in their papers 
\item 
the suggestion that comparison tables among learners, presented in the literature, would benefit from being expanded and recalculated according to performance maps to provide more insights to the reader looking for the best learner/configuration when dealing with a specific data set.  
\item 
the observation that performance maps fit nicely in the scope of the No Free Lunch Theorem (NFL) \cite{NFL1997}. The NFL theorem states that no learning algorithms can outperform all the others over all data sets.  
Our proposal makes explicit that changing parameter settings of a learning algorithm produces a different learner which usually has different a performance. 
\end{enumerate}

The paper is organized as follows: in Section 2, we summarize the standard procedure to compare learning systems, in Section 3, we introduce the learning systems and the meta-optimizers used in the study, in Section 4 and 5, we discuss their parameter spaces, Section 6 describes the data sets used in the experiments, Section 7 reports the experimental study, and finally some conclusions close the paper.

\section{State of the art in comparing learning algorithms}
The standard procedure to compare learning algorithms consists in contrasting their performances on several data sets. It must be added that the comparison is done after an ad hoc selection of the better performing parameter settings for the learners. Usually manually discovered by running some trial tests.  

Traditional performance measures include: accuracy, error rate, $R²$, etc. Their values are generally determined by using a statistical methodology called n-fold cross validation (usually 5 or 10 folds are selected) on the whole available data in order to determine a performance interval (mean $\pm$ standard deviation) with known statistical confidence \cite{Refaeilzadeh2009crossvalidation,Stone1974crossvalidation}.

Because performance measures reduce to a single value the whole learner's behavior,
they may potentially miss important aspects of the underlying learning process like, for instance, the distribution of the performances over the parameter space of the learner.

In addition to traditional performance measures, other methodologies exist to evaluate a learner's performance. For instance: the
Area Under the ROC Curve (AUC) \cite{Bradley1997AUCROC} or the rolling cross validation \cite{Racine2000rollingCV,Bergmeir2012rollingCV,Neri201286}. 
AUC is applicable to any classifier producing a score for each case, but less appropriate for discrete classifiers like decision trees.
Rolling cross validation is only applicable to specific data types like time series or data streams \cite{Racine2000rollingCV,Bergmeir2012rollingCV,Neri201286}. 
In fact, more recent performance measures are not generally applicable across learners or data types.

We then believe that, when learners need to be compared, the information provided by the above performance measures could be enhanced by including some insights about the distribution of performances on the learners' parameter spaces.  
The latter information would allow, for instance, to take into account the probability of achieving a high performance by randomly selecting a parameter from the learner's parameter space  with uniform probability. Thus providing a measure of confidence or stability in the best performance achieved in the learning context under study.  

\subsection{Our proposal: comparing learning algorithms with performances maps and their HP(k) values \label{hpsection}}
This study proposes to compare learning algorithms by confronting their \emph{performances maps} and 
their HP(k) values.
As said, given a learning context $LC$, its performance map $Pmap(LC)$ is the collection of pairs
  \mbox{< s, $L(s)$>}, with $s$ in $MOPS$, and $L(s)$ as the performance of $L$ run with settings $s$.
From $Pmap(LC)$, it is very simple to determine its best performance $best(LC)$ (the map's maximum).

The High Performance function of a map $HP_{Pmap(LC)}(k)$ is defined 
as the ratio between the number of parameter settings in $MOPS$ producing a 
performance with distance $k$ from $best(LC)$, and the cardinality of $MOPS$, as in eq. (\ref{eqhp}). 
\begin{equation}
HP_{Pmap(LC)}(k) = \frac{|\{p | p \in MOPS \land \; L(p) \geq best(LC) * (1-k) \}|}{|MOPS|}
\label{eqhp}
\end{equation}
\noindent
where $p \in MOPS, 0 < k < 1$, and $L(p)$ is the performance observed by running $L$ with parameter settings p on the data D.
In the following, we will use $HP_{LC}(k)$, or simply $HP(k)$ when the learning context is clear, as shorthand for $HP_{Pmap(LC)}(k)$.

$HP_{LC}(k)$ also represents the fraction of the map area above a certain performance level ($best(LC) * (1-k)$) over the whole map extension.  
And, from another point of view, $HP_{LC}(k)$ is an estimate of the cumulative distribution function  $Prob_{LC}(X > best(LC) * (1-k))$, where $X$ is $L(s)$ and $s$ is randomly taken from $MOPS$ with uniform distribution. 

We will show, in the experimental session, the values of $HP_{LC}(k)$ for several learning contexts.

\section{Learners and meta optimization methods}
As said, the aim of our work is to compare learners across learning contexts by using performance maps.
In order to practically show how our proposal works, we selected two learners and two meta-optimization methods so that we were able to present full set of experiments.

Decision Trees (DT) \cite{quinlan:c45} and Support Vector Machines (SVM) \cite{vapnikSVM1995} are selected as learners because they internally represent knowledge in very different way, thus demonstrating the general applicability of our  methodology. And as meta-optimization methods, we selected Grid Search, which consists in the exhaustive enumeration of a input parameter space, and Simple Genetic Algorithm (SGA) \cite{goldberg:book,neriGAintrusions}, in order to account for the case of partial search of the input parameter space, and the ensuing partial performance map. We note that one can choose to build a partial performance map as it has a lower computational cost than a complete one. 
The pseudo-code for the used SGA and Grid Search can be found in Appendix \ref{appSGAgs}.

\section{The Parameter Spaces for the selected Learners \label{paramspaces}}
The chosen parameter spaces for  DT and  SVM are shown in Tables \ref{parDT} and \ref{parSVM}. These are the parameter spaces searched by the meta-optimizer. 

In the case of DT, the parameters that mostly affects its results have been identified in: minimum impurity decrease (decrease of a node's impurity to allow for a node splitting), minimum samples (the minimum number of samples required to split an internal node), and max depth (the maximum allowed depth of the tree). 
The parameter space for DT contains combinations of values for the three selected parameters.
Similarly, for  SVM, the chosen parameters are gamma, kernel, and C value, which affect the types of hyperplanes to be used and their boundary positions (margin distance). Again the combination of values for these three parameters define the parameter space for SVM.

It is important to note that our methodology is not limited by the number of parameters used to define a parameter space. In this experimentation, we define the parameter spaces with only three parameters per learner simply because this choice will allow to draw 3-dimensional representation of the performance maps build in the experiments. Thus facilitating the understanding of our work. If we had used more parameters it would have been difficult to show the results in a graphical form.

 \begin{table*}[htb]
\caption{Value ranges for the selected parameters of DT.}
\label{parDT}
\begin{tabular}{| c | c | c | c | c |}
\hline 
Learner &		Min Impurity	& Min Samples  & Max Depth & Timeout (secs)  \\
\hline
DT  &		\{i/10 for i = 0 to 6\} & \{i for i = 2 to 150 step 10\} & \{i for i = 1 to 160 step 10\} & 40 \\
\hline   
\end{tabular}
\end{table*}

\begin{table*}[htb]
\caption{Value ranges for the selected parameters of SVM.}
\label{parSVM}
\begin{tabular}{| c | c | c | c | c |}
\hline   
Learner &			Gamma	& Kernel  & C value & Timeout (secs)  \\
\hline  
	SVM &	   scale & linear  & \{i/100 for i = 1 to 200 step 20\} $\bigcup$ & 40 \\
		&		auto & poly    &  \{i for i = 2 to 200 step 20\} & \\
		&		     & rbf     &  & \\
		&		  	 & sigmoid &  & \\
\hline  
\end{tabular}
\end{table*}
The Timeout columns in the tables report the maximum number of seconds an experiment will run before timing out. As an anticipation, an experiment consists in performing several 10 fold cross validations of the selected learner on the available data in order to meta-optimize it.

Using a timeout is necessary for some data sets and learners given the long run time required. In this study, the time out is particularly needed when SVM is applied to the Pima Indians Diabetes and Abalone data sets which may requires more than 30 minutes for each experiment. Resulting in a full experimentation running for several hours. The use of timeouts does not affect our comparison methodology though it may produce approximate performance maps. We denote a timeout experiment with a negative value equal to -0.2 on a performance map.

\section{Parameter settings for the meta-optimization methods }
In the case of Grid Search, no parameters affects its behavior because all points in the given parameter space are evaluated.

In the case of SGA, instead, it is known that the population size and the maximum number of generations can deeply affect the result found by a genetic algorithm.
Here is why, in order to find the best parameter settings for the SGA, 
we meta-optimized the SGA by using a Grid Search applied to the following parameter ranges: 
population size (30, 50, 80), max number of generations (30, 50, 80), crossover probability (0.5, 0.7, 0.9), and learner (DT or SVM). 

As performance measure, we were interested in the genetic algorithm discovering a parameter settings
performing as close as possible to the best performance discovered by Grid Search when used as a meta-optimizer in the learning contexts. Also by using the lowest possible population size and max generations.

The found parameter settings for SGA are: population size equal to 50, max number of generations equal to 50, and crossover probability equal to 0.9. 
The fact that genetic algorithms, in general, are robust learners makes it quite easy to find one of the many suitable parameter settings \cite{neriGAintrusions,neriPIRR}.

We kept the remaining parameters of SGA to their default values as set in the python library GeneticAlgorithm \\
(https://pypi.org/project/geneticalgorithm/) from which we built the SGA used in this study.

\section {Data set descriptions} 
To perform the experiments in our study, we selected four data sets with varying characteristics from the UCI Machine Learning repository: 
\begin{enumerate}
	\item
	Mushrooms - 8124 instances, 22 attributes (categorical), classification task: to predict if a mushrooms is either edible or poisonous from some physical characteristics \cite{Schlimmer1987}.
	\item
	Pima Indians Diabetes - 769 instances, 8 attributes (categorical), classification task: to predict if the patients has or not diabetes based on some diagnostic measurements. Source: https://www.kaggle.com/uciml/pima-indians-diabetes-database
	\item
	Congressional Voting Records - 435 instances, 16 attributes (categorical), classification task: predicting Republican or Democratic membership from vote record \cite{Schlimmer1987}.
	\item 
	Abalone - 4177 instances, 8 attributes (categorical, integer, real), regression task: predicting the age of abalone (a marine snail) from its physical measurements \cite{Waugh1995}. 
\end{enumerate}
An open research question is if the proposed methodology needs to be extended when different data types like for instance financial time series \cite{Neri2012admi, Neri2010versSim, Neri2012202, Neri2011lncsABM} or unusual domains are considered \cite{MagarinoNeriPlaza2019}. 

\section {Experimental analysis} 
As experimental platform, we implemented the code in Python 3.8, making use of SciKit Learn \cite{scikit-learn}, and used a Dell XPS 13, with Intel CPU I7, 7th gen, and 16 GB RAM, as hardware. 
We used the implementation of DT and SVM as provided in python's SciKit Learn library, and directly implemented the meta-optimization algorithms (Grid search and SGA) whose pseudo code can be found in 
Appendix \ref{appSGAgs}.

Given a learning context $LC$, an experiment consists in using the meta-optimizer $M$ to find the best performing parameter settings for the learner $L$. Each parameter settings evaluated by $M$
requires performing a 10 fold cross validation in order to ensure the correct measurement of $L$'s performance.

In Table \ref{expres}, all the experiments performed are reported with their best performances across the 16 learning contexts considered. Performances are measured with the accuracy measure for classification tasks,
and with the coefficient of determination $R^2$ for the regression task (Abalone data set). 
The time column shows the time to run a complete experiment.      
The following findings appear from Table \ref{expres}:
\begin{enumerate}
	\item some learning contexts do not admit for a perfect solution; 	
	\item learning contexts with DT and SVM display similar performances except in the case of the Abalone data set;
	\item learning contexts with Grid Search, as a meta-optimizer, usually takes longer than SGA. This is reasonable  because Grid Search has to evaluate all settings for $L$, whereas SGA will consider only some of them.
	\item learning contexts with DT usually run in less time than those exploiting SVM.  
\end{enumerate}
The structure of these findings follows the standard used in the literature to assess learning systems.

In this study, however, we want to augment the way machine learning systems are compared
by including also information from \emph{performance maps} and their \emph{high performance values}. 

\begin{table*}[htb]
	\caption{Meta-optimization of learners in several learning contexts. }
	\label{expres}
	\begin{tabular}{cccccc}
		Data set   & Learner and Meta Optimization & Best Accuracy/$R^{2}$ & Std & Evaluated points & Time     \\
		\hline
		Mushrooms & DT - Grid & 1.0     & 0.0 &	1440	& 197.45  \\
		& DT - SGA    		& 1.0       & 0.0 & 	49	& 6.70   \\
		& SVM - Grid 		& 1.0       & 0.0 &	160		& 1000.25 \\
		& SVM - SGA   		& 1.0       & 0.0 & 	47	& 320.30 \\			
		\hline
		Congr. Votes   & DT - Grid  & 0.96      & 0.03 &	1440	& 18.08  \\
		& DT - SGA    				& 0.96      & 0.03 & 	272	& 5.28   \\
		& SVM - Grid 				& 0.97      & 0.02 &	160	& 5.50 \\
		& SVM - SGA   				& 0.96      & 0.02  & 	129	& 6.11 \\	
		\hline
		Diabetes & DT - Grid  & 0.75   		& 0.04 & 1440  &  62.53  \\
		& DT - SGA    		  & 0.75     & 0.04 & 	241	 &	12.50   \\
		& SVM - Grid 		  & 0.76     & 0.04 & 	160	 &	2312.26 \\
		& SVM - SGA   		  & 0.76     & 0.04 &	122	 &	2127.17 \\
		\hline
		Abalone   & DT - Grid   & 0.49     & 0.02 & 1440	& 133.13  \\
		($R^{2}$) & DT - SGA     & 0.49     & 0.02 & 291		& 35.22   \\
		& SVM - Grid 			& 0.56     & 0.02 &	160		& 1512.06 \\
		& SVM - SGA   			& 0.56     & 0.02 &	109		& 1068.36 \\
		\hline 
	\end{tabular}
\end{table*}
  
\subsection{Performance maps}  
We recall that a performance map $Pmap(LC)$ for a learning context $LC$ 
is the set of pairs $< s , L(s)>$, where $s$ is 
a parameter settings in $MOPS$ and $L(s)$ is the performance obtained by running $L$ with settings $s$. 

\emph{Because a performance map, for a learning context, shows the distribution of performances for the associated learner over (part of) its parameter space, 
it then provides information about 
how frequent high performing parameter settings are.
Also it shows the specific value ranges for those high performing  settings.
Thus, in addition, a performance map provides  an insight about how robust the associated learner is to changes to its settings.
}

Building a performance map then could be particularly useful when selecting
a learner for some novel data, because it provides information on the robustness of the learner
when different configurations are used, a situation which is bound to happen in real world usage of a learning system.
 
Here is why we believe that comparing learner by using performance maps provides more insights
than the use of a single valued performance measure as traditionally done in the literature.

\begin{figure*}[htb]
	\centering
	\subfloat[]
	{\includegraphics[height=1.5in,width=1.5in]{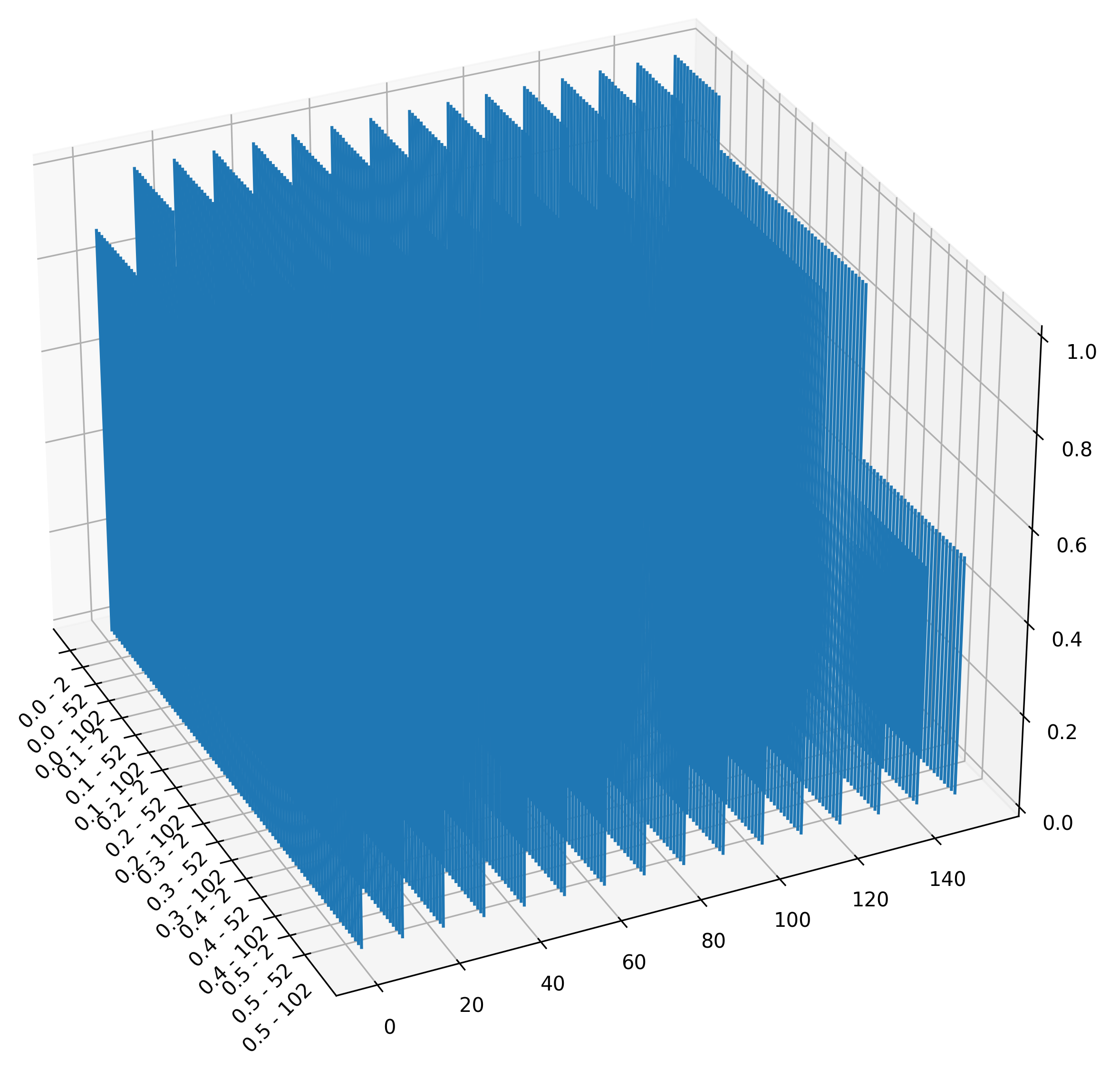}\label{fig:pict1}} 
	\hfil 
	\subfloat[]
	{\includegraphics[height=1.5in,width=1.5in]{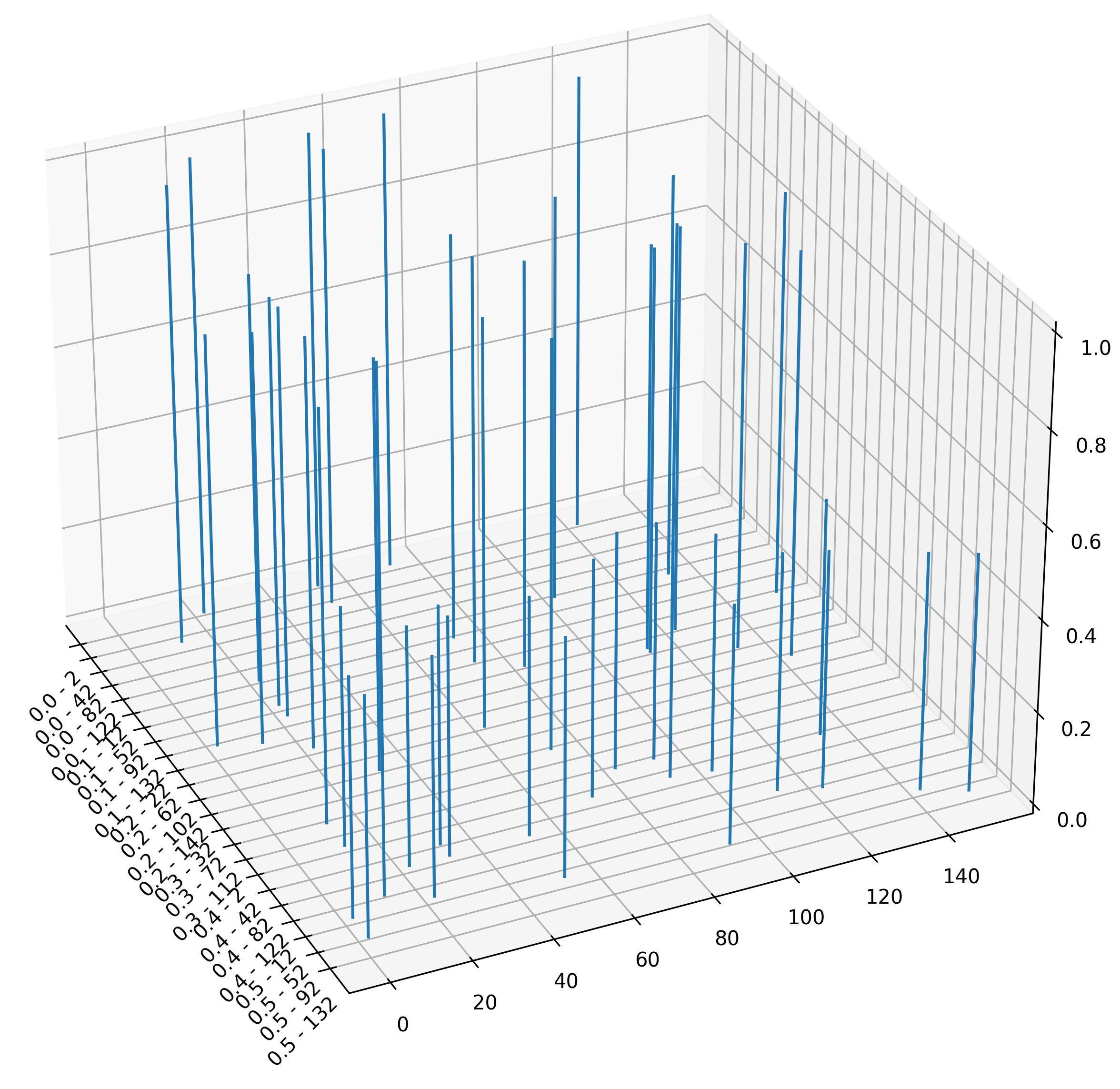}\label{fig:pict2}} 
	\hfil
	\subfloat[]
	{\includegraphics[height=1.5in,width=1.5in]{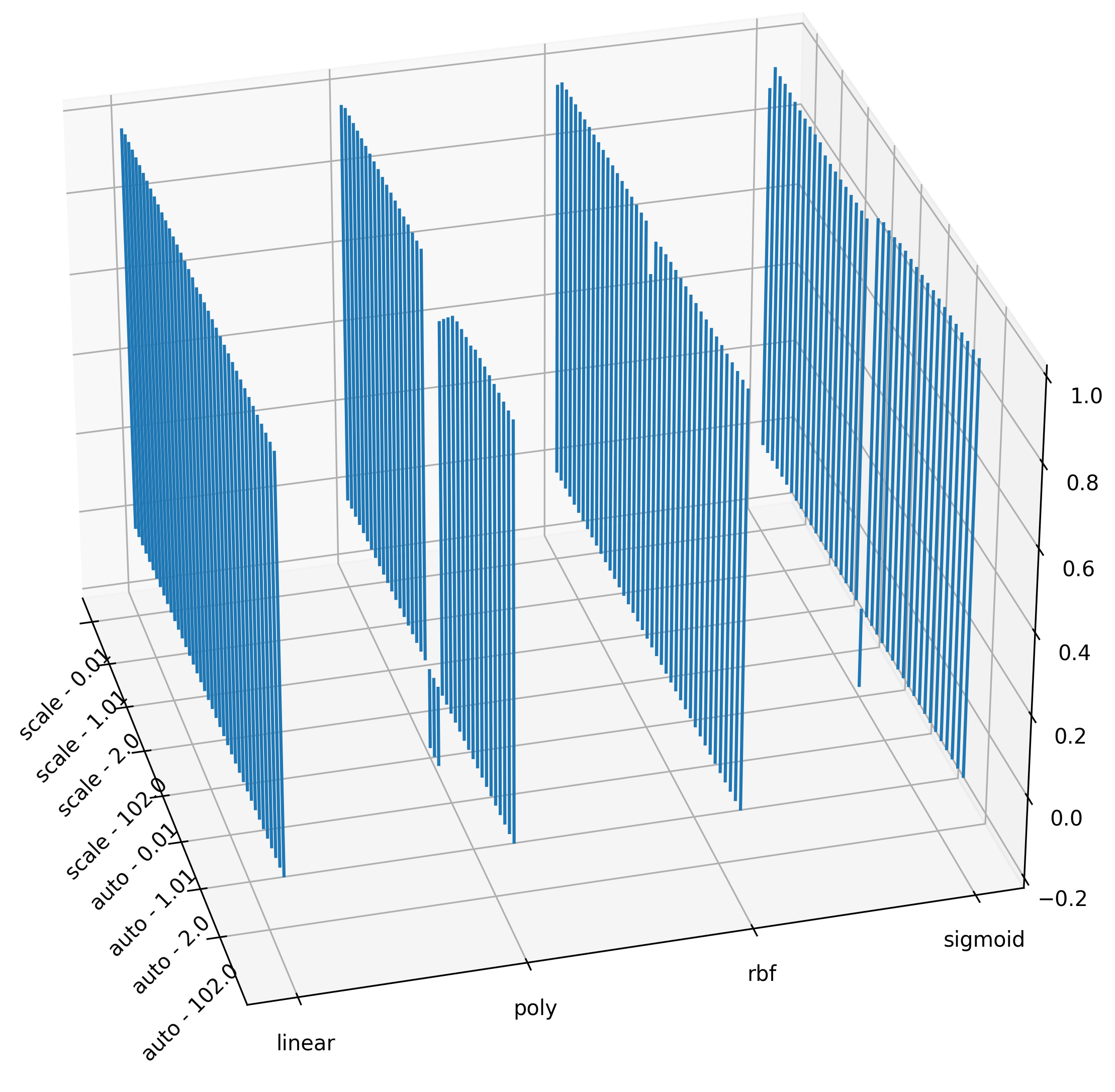}\label{fig:pict3}} 
	\subfloat[]
	{\includegraphics[height=1.5in,width=1.5in]{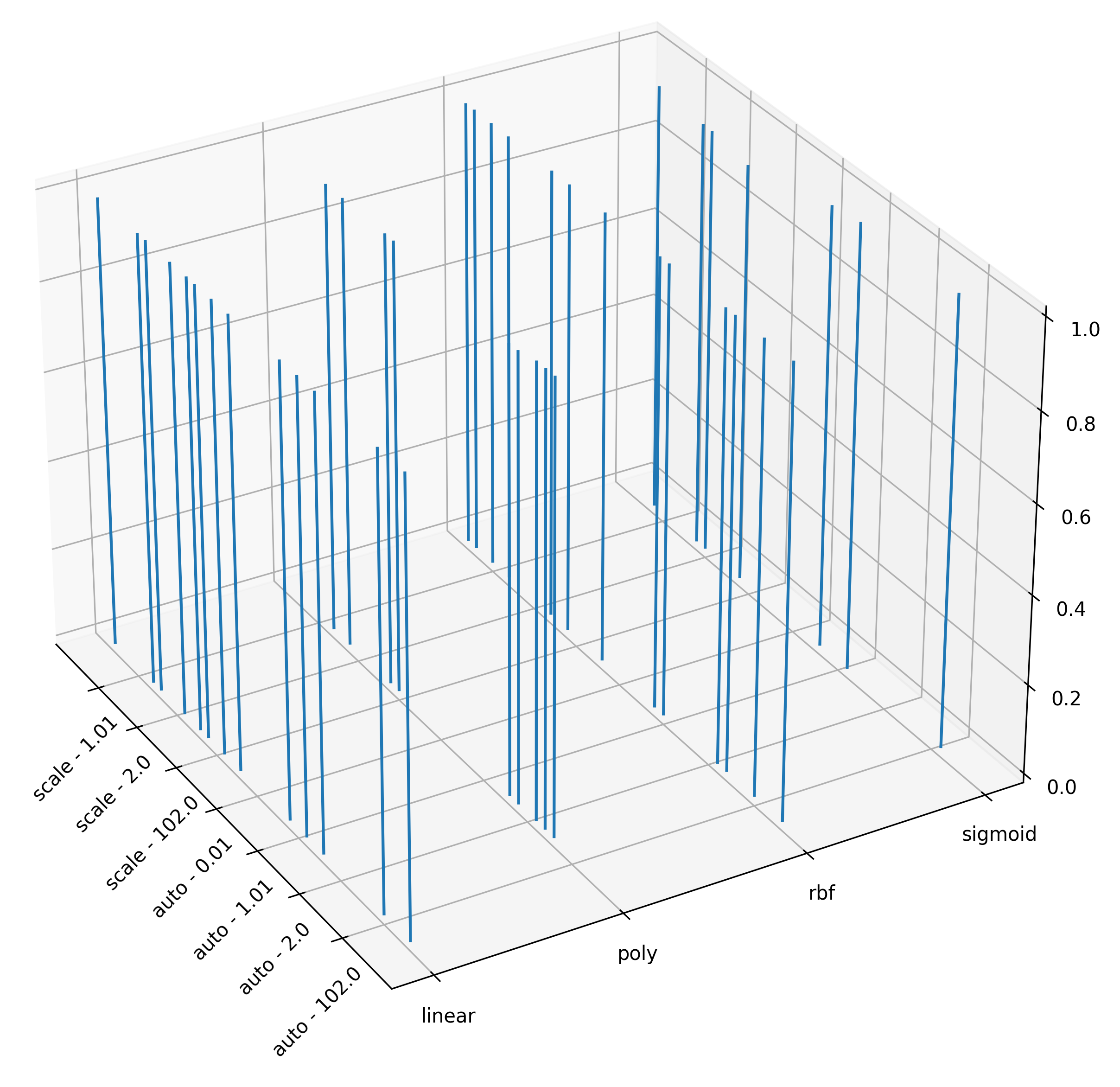}\label{fig:pict4}} 
	\caption{Performance maps for the mushrooms data set. In the cases of <DT, Grid search> (a), 
	<DT, SGA> (b), <SVM, Grid search> (c), and <SVM, SGA> (d). \label{res1}}
\end{figure*}

\begin{figure*}[htb]
	\centering
	\subfloat[]
	{\includegraphics[height=1.5in,width=1.5in]{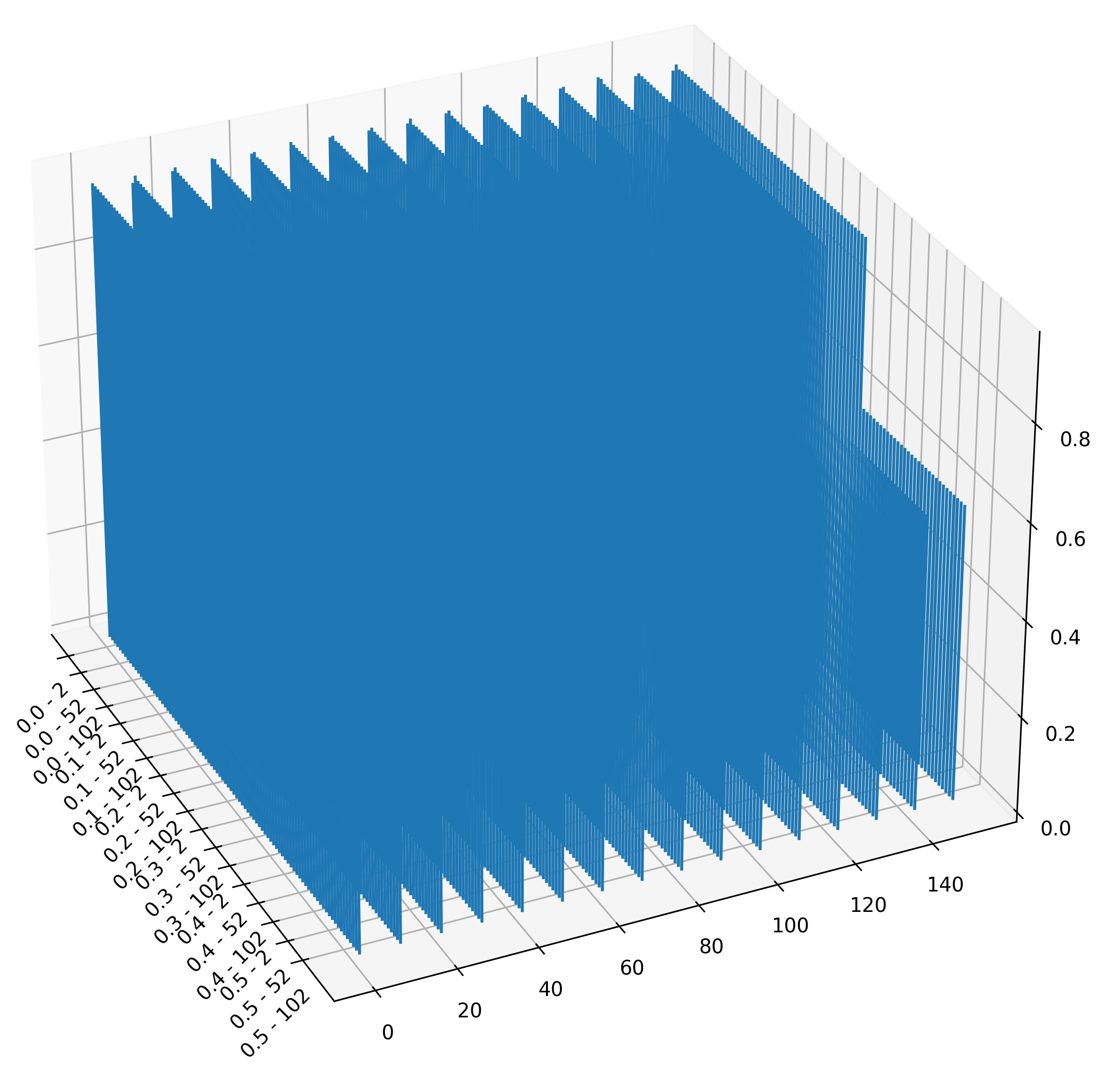}\label{fig:pict1}} 
	\hfil 
	\subfloat[]
	{\includegraphics[height=1.5in,width=1.5in]{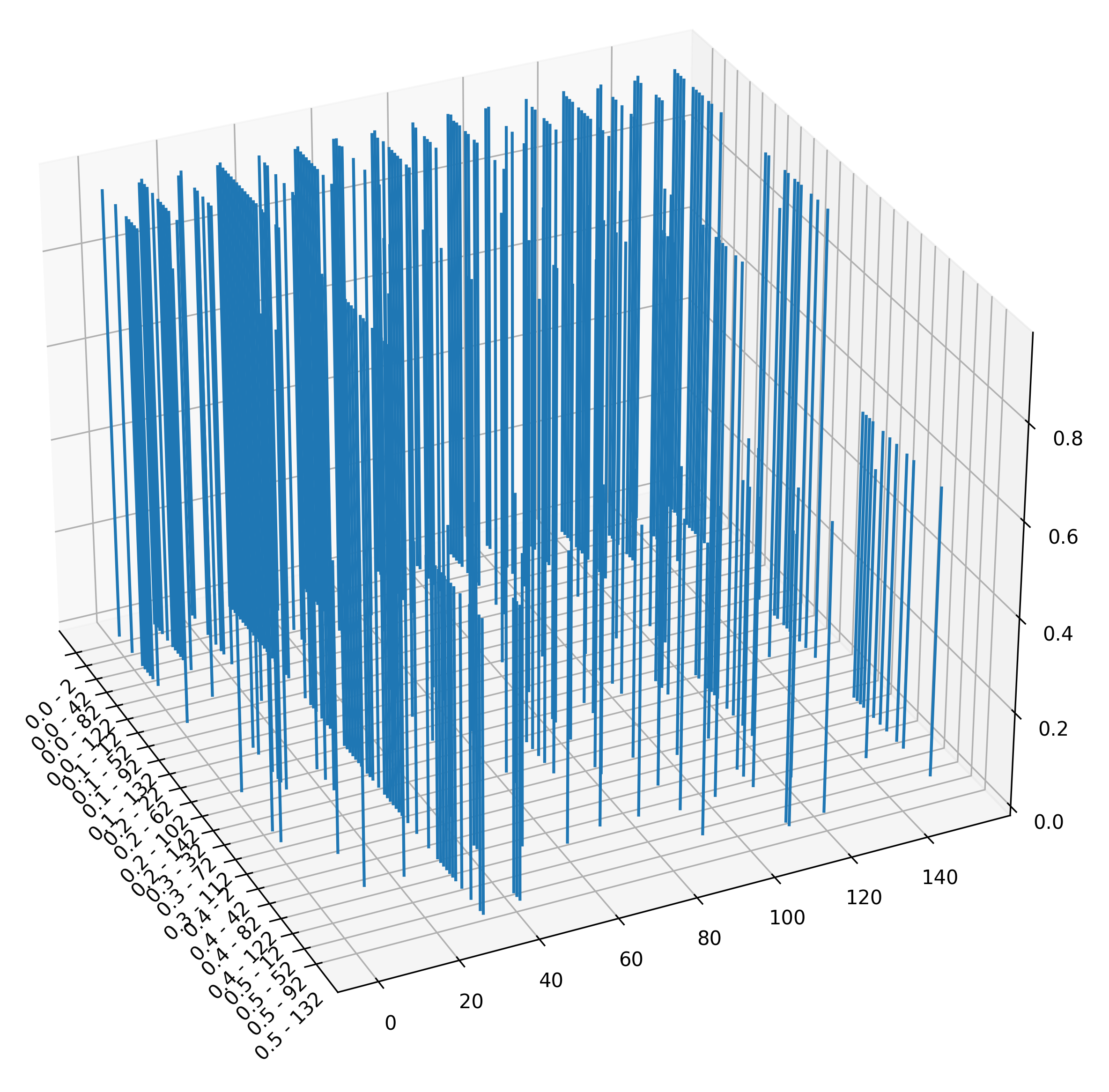}\label{fig:pict2}} 
	\hfil
	\subfloat[]
	{\includegraphics[height=1.5in,width=1.5in]{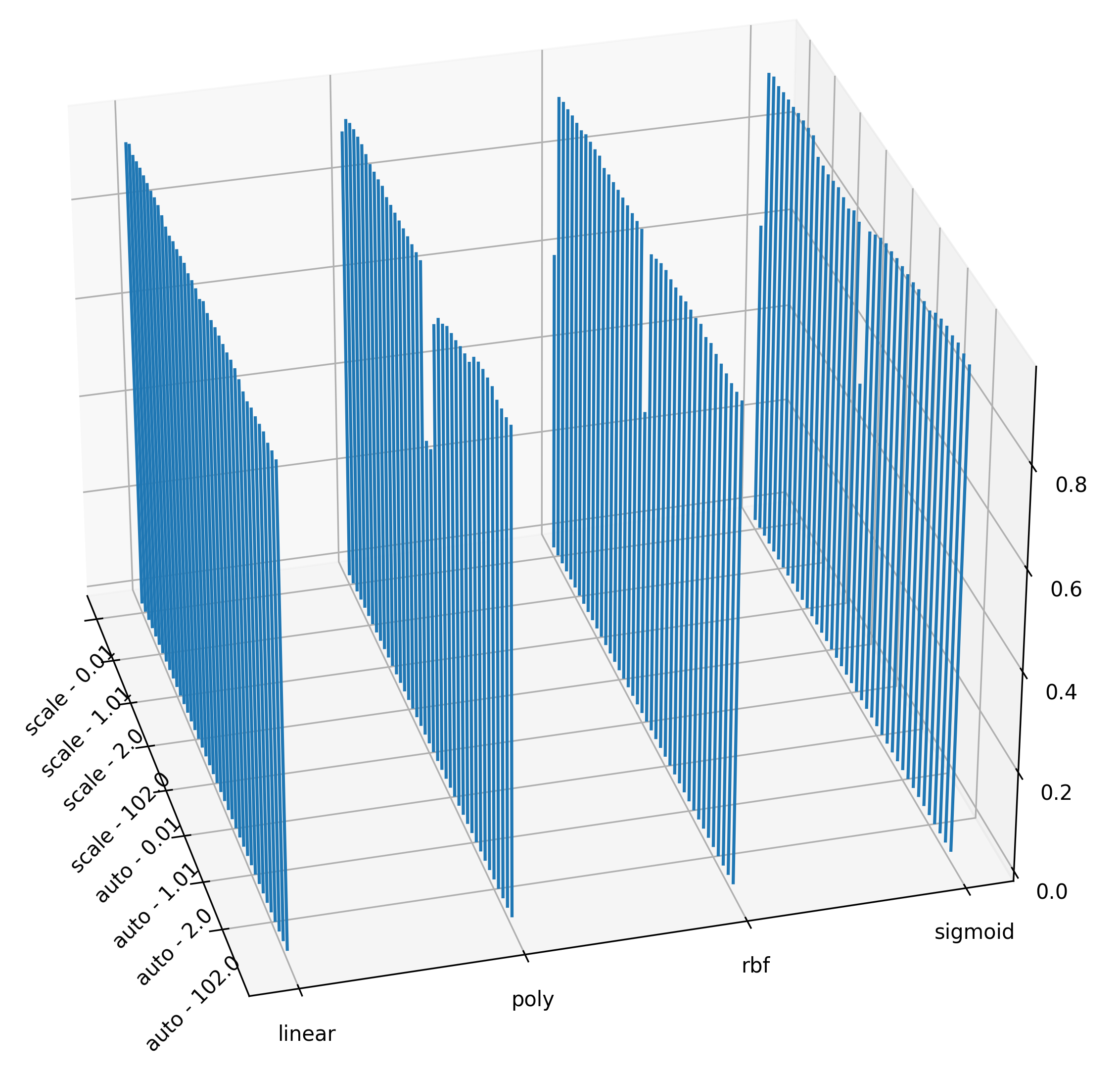}\label{fig:pict3}} 
	\subfloat[]
	{\includegraphics[height=1.5in,width=1.5in]{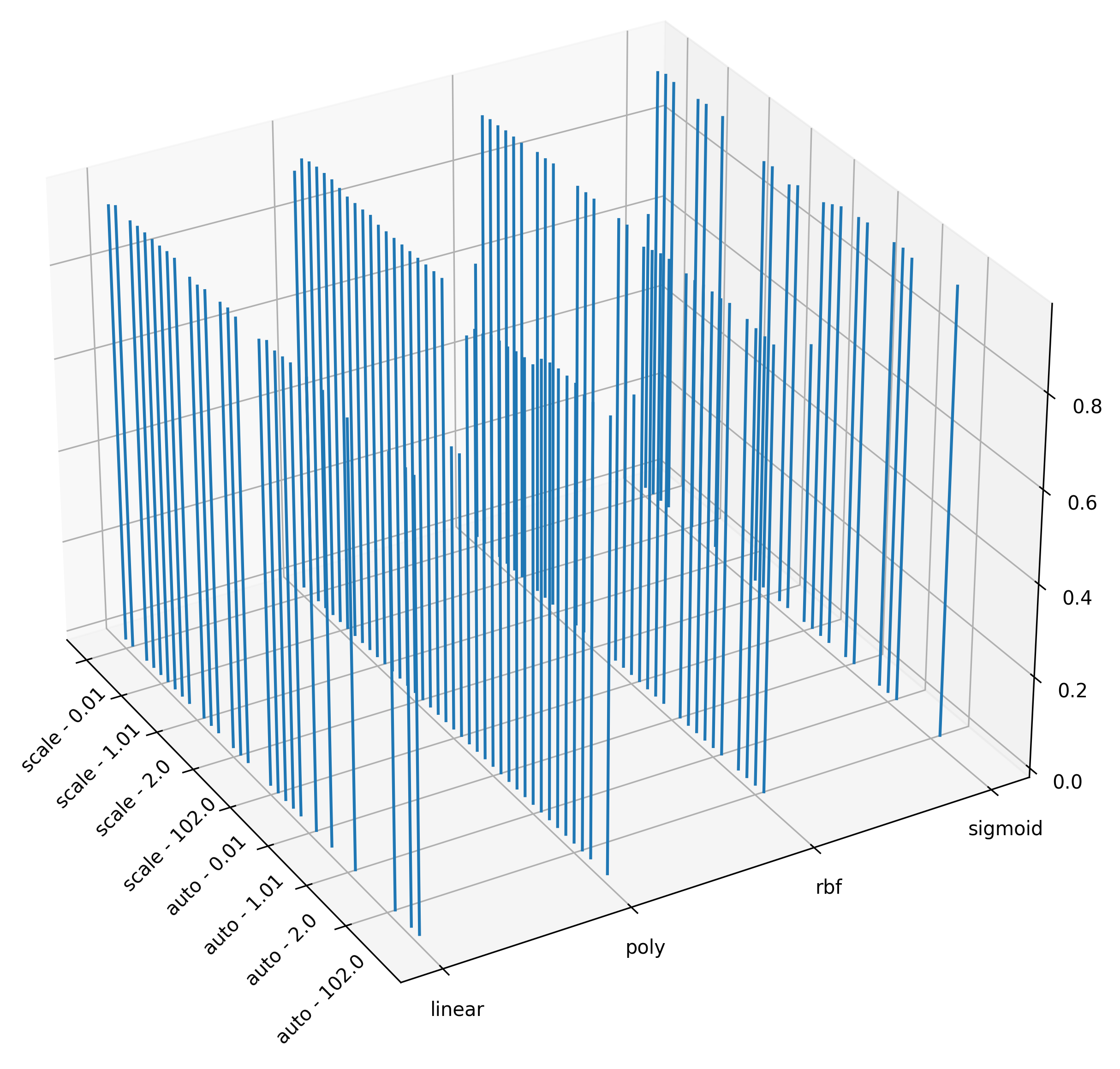}\label{fig:pict4}} 
	\caption{Performance maps for the congressional voting records data set. In the cases of <DT, Grid search> (a), <DT, SGA> (b), <SVM, Grid search> (c), and <SVM, SGA> (d). \label{res2}}
\end{figure*}

\begin{figure*}[htb]
	\centering
	\subfloat[]
	{\includegraphics[height=1.5in,width=1.5in]{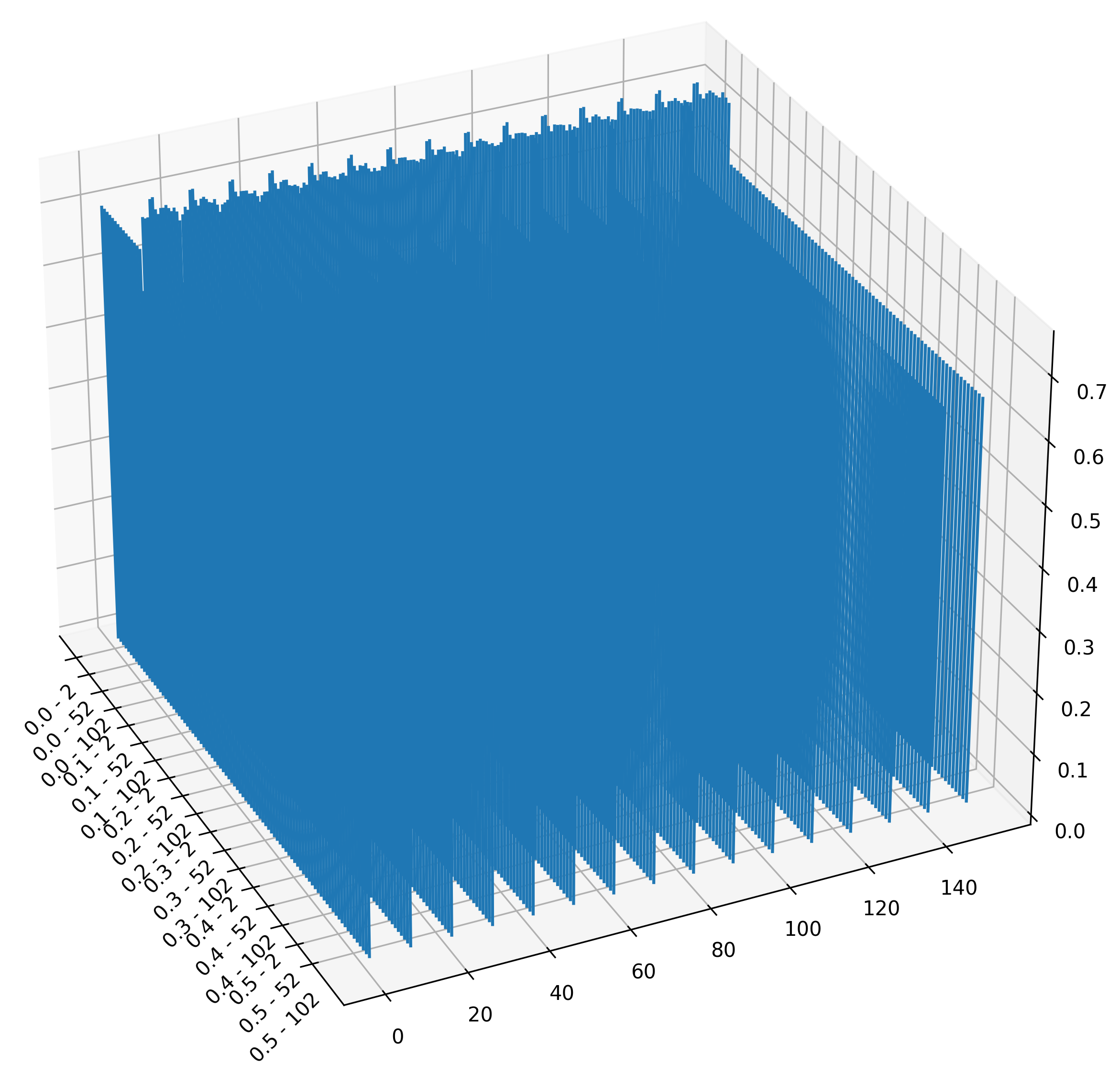}\label{fig:pict1}} 
	\hfil 
	\subfloat[]
	{\includegraphics[height=1.5in,width=1.5in]{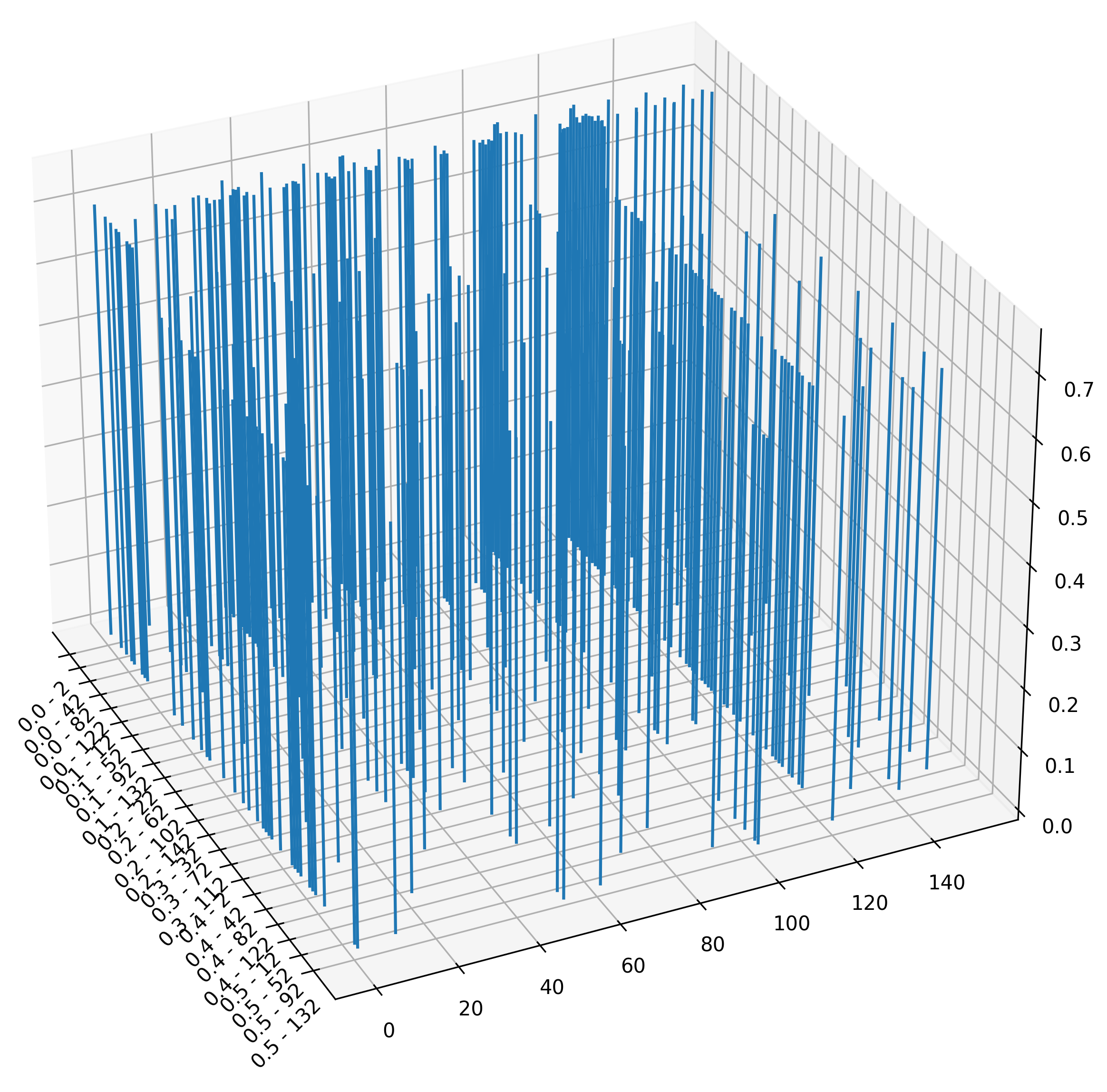}\label{fig:pict2}} 
	\hfil
	\subfloat[]
	{\includegraphics[height=1.5in,width=1.5in]{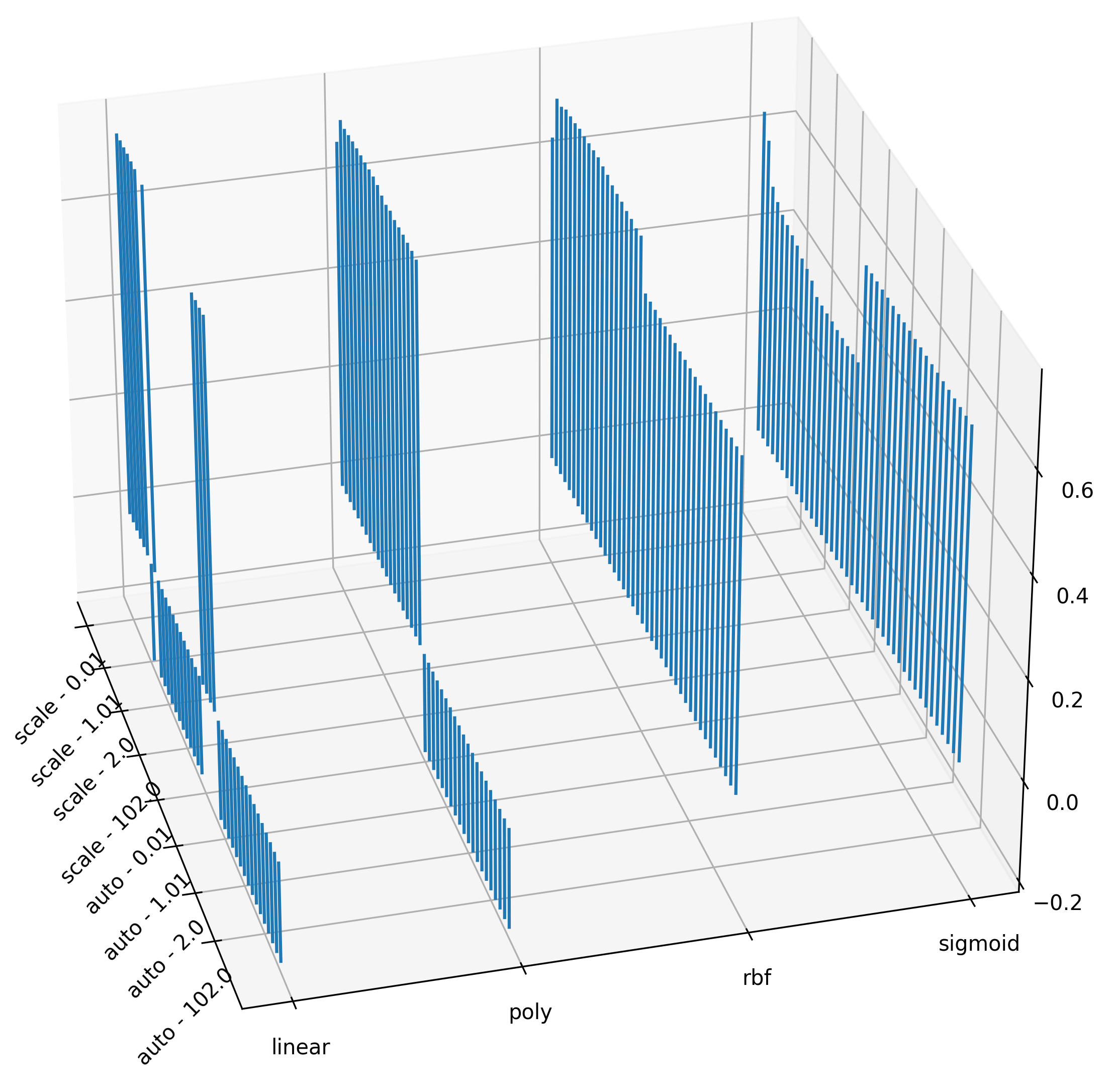}\label{fig:pict3}} 
	\subfloat[]
	{\includegraphics[height=1.5in,width=1.5in]{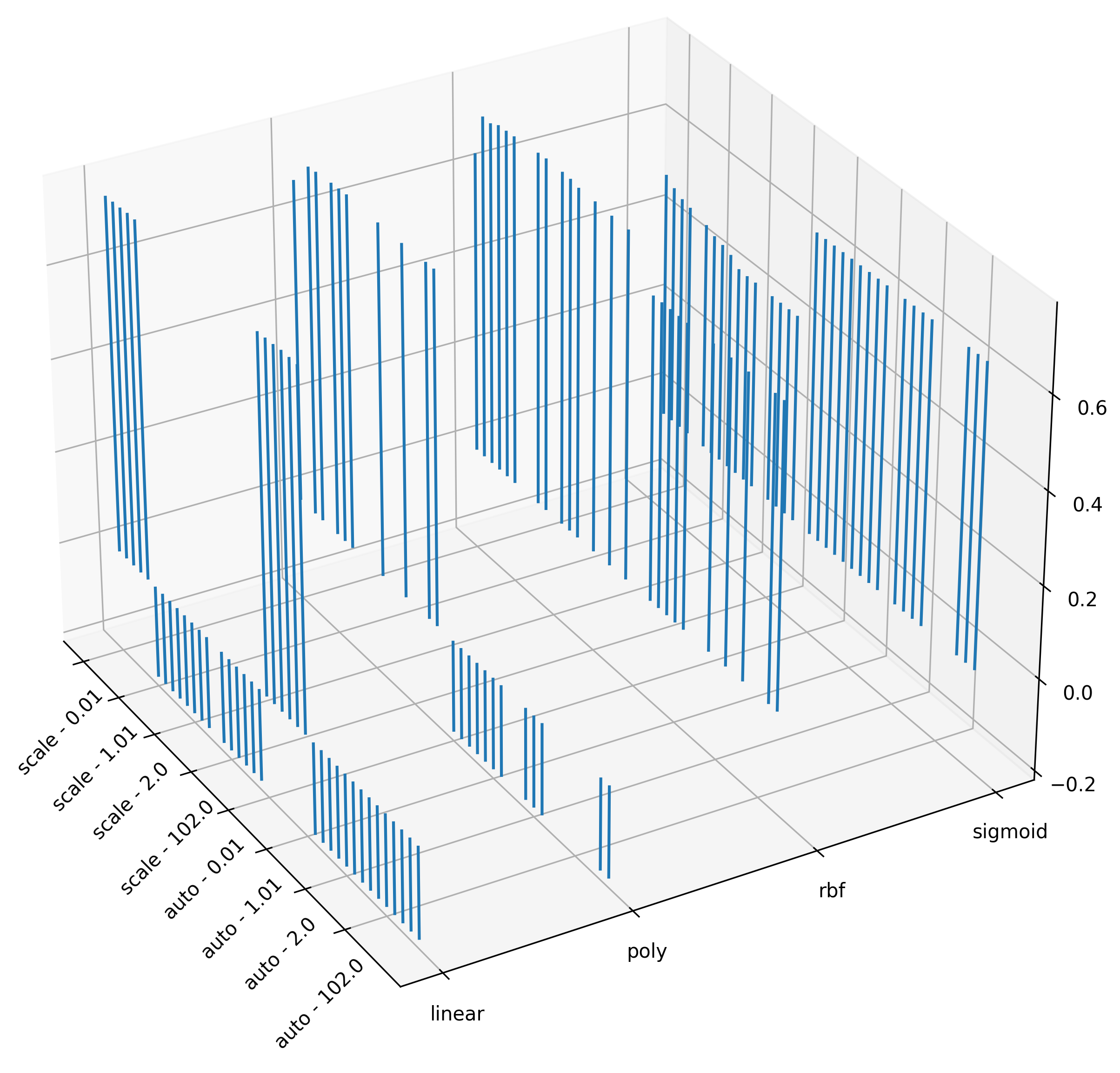}\label{fig:pict4}} 
	\caption{Performance maps for the pima indians data set. In the cases of <DT, Grid search> (a), 
		<DT, SGA> (b), <SVM, Grid search> (c), and <SVM, SGA> (d).\label{res3}}
\end{figure*}

\begin{figure*}[htb]
	\centering
	\subfloat[] 
	{\includegraphics[height=1.5in,width=1.5in]{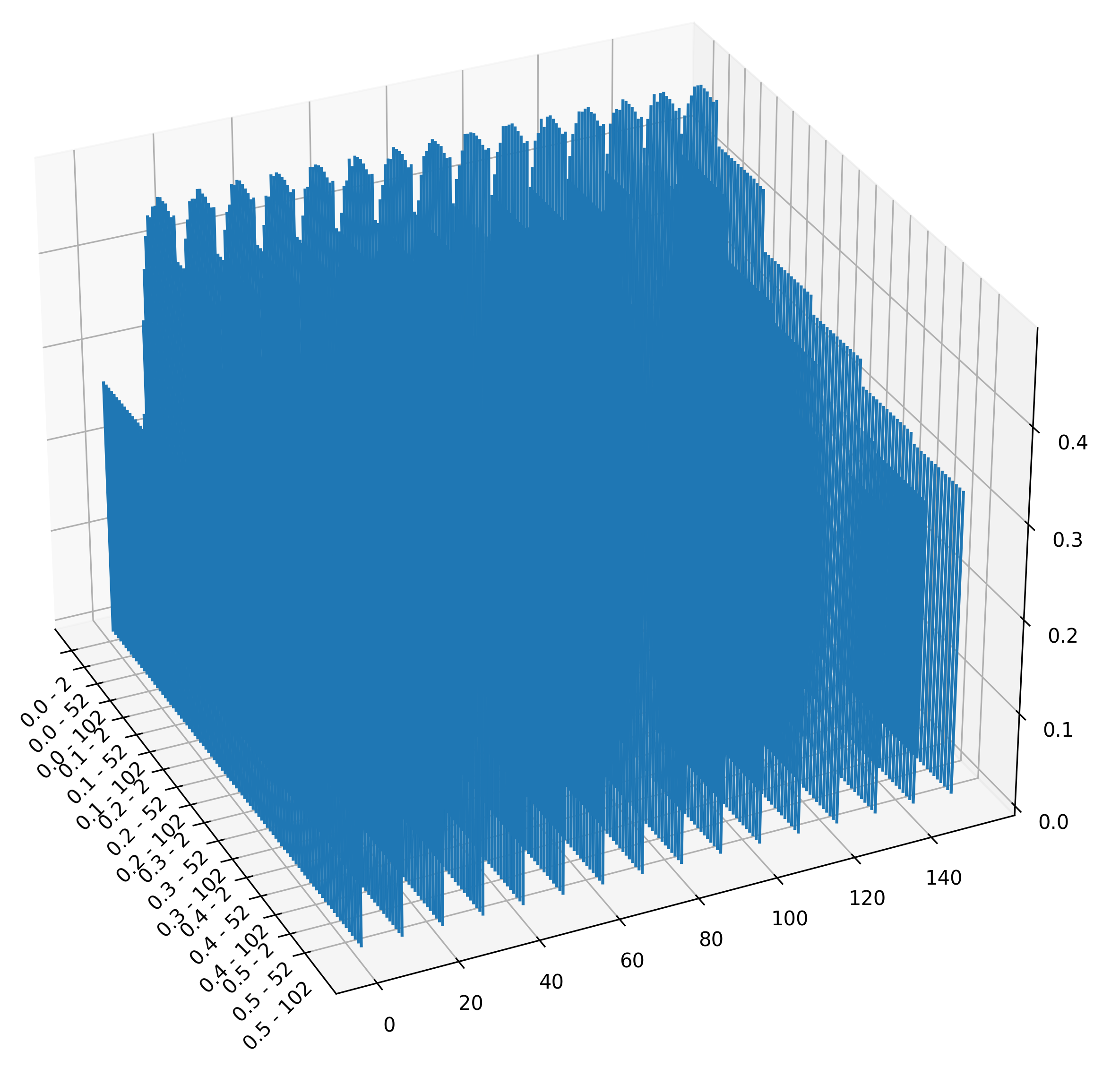}\label{fig:pict1}} 
	\hfil 
	\subfloat[]
	{\includegraphics[height=1.5in,width=1.5in]{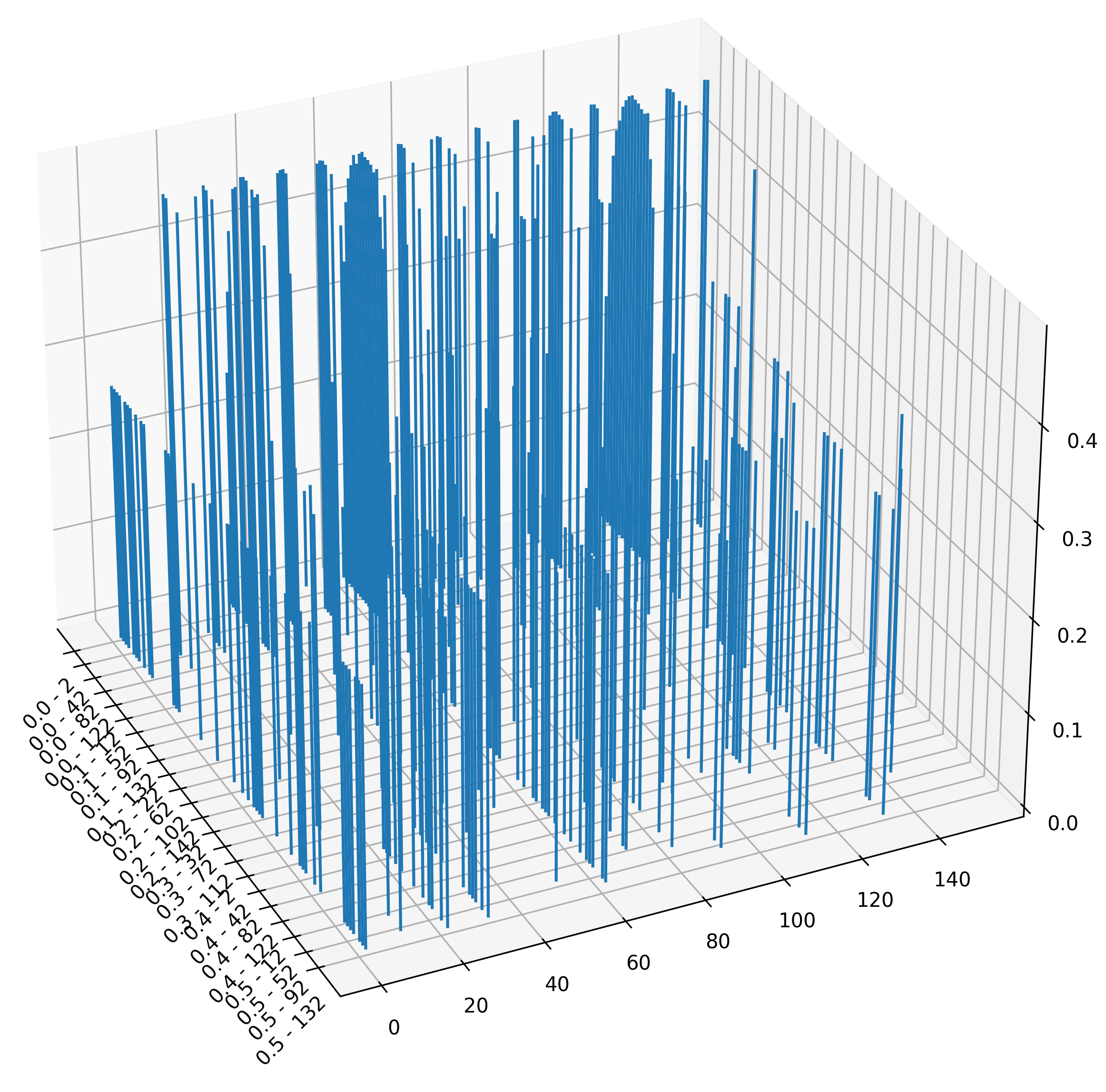}\label{fig:pict2}} 
	\hfil
	\subfloat[]
	{\includegraphics[height=1.5in,width=1.5in]{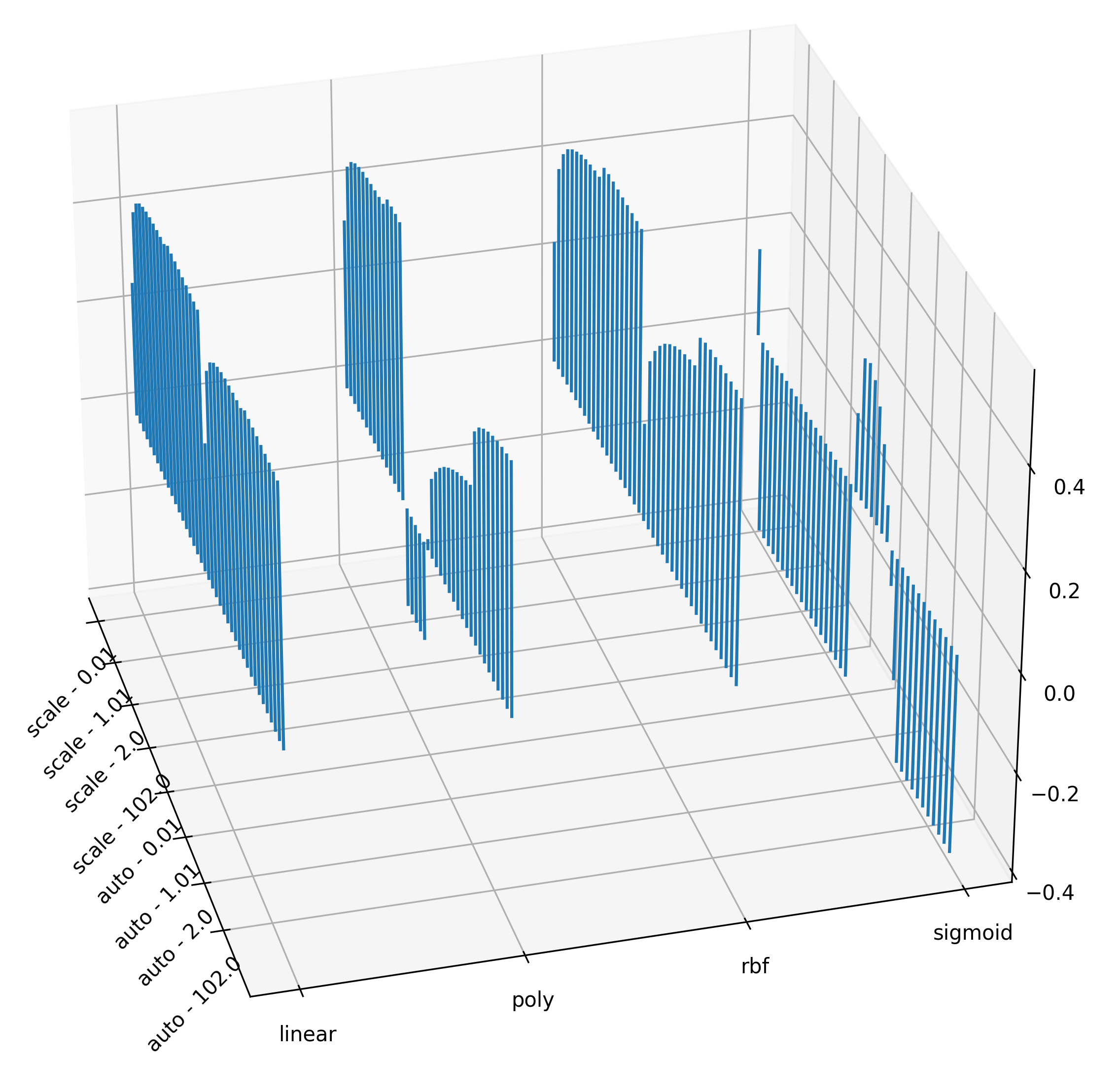}\label{fig:pict3}} 
	\subfloat[]
	{\includegraphics[height=1.5in,width=1.5in]{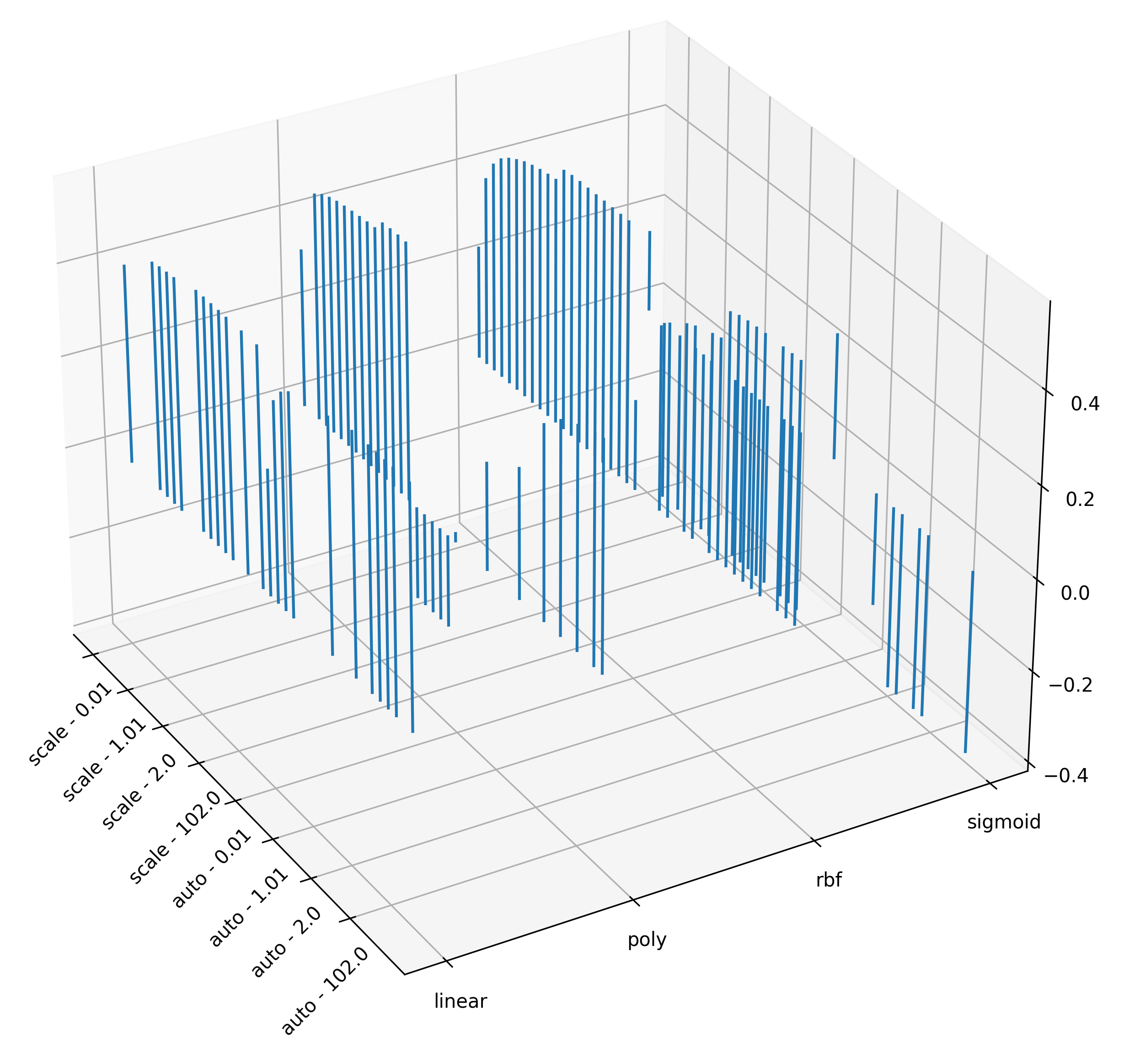}\label{fig:pict4}} 
	\caption{Performance maps for the abalone data set. In the cases of <DT, Grid search> (a), 
		<DT, SGA> (b), <SVM, Grid search> (c), and <SVM, SGA> (d). Note that the highest points in (c) for 'poly' and 'rbf' reaches the value of 0.56. Perspective makes them appear to be lower. \label{res4}}
\end{figure*}

Figures \ref{res1}, \ref{res2}, \ref{res3}, \ref{res4} show the \emph{performance maps} for the learning contexts of Table \ref{expres}\footnote{We projected two parameters on the X axis for creating the 3-d graphs. 
	In particular, we projected 'min impurity' and 'min samples' on the X axis and 'max depth' on the Y axis
	for DT. Then the label '0.1 - 20' on the X axis has to be interpreted as 'min impurity' = 0.2 and 'min samples = 20'. Instead, for SVM, we projected  'gamma' and 'C value' on the X axis and  'kernel' on the Y axis.}. 
They peruse makes explicit that:
\begin{enumerate}
	\item if we consider all learning contexts, DT performs better in a region of the parameter space where 'min impurity' is close to 0, 'min sample' is below 50 and 'max depth' is above 20. When increasing the 'min impurity' value above 0.2, the performance decreases abruptly and significantly
	\item if we consider all learning contexts, SVM performs better in a region or the parameter space where 'gamma' is equal to 'scale', 'C-value' is lower than 1.0, and 'kernel' is 'poly', 'rbf' or 'linear'
	\item however, if we are interested in a specific learner and data, the performance map shows the locations of the highest performing parameter settings and it displays how these regions varies in location and extensions across the parameter space
	\item performance maps do not need to be complete to be useful. Completeness may require a high computational cost to achieve. Indeed, even partial performance maps are very helpful in selecting high performing parameter settings over just a blind selection of the same done by manually undertaking trial runs. 
	Comparing performance maps using Grid Search with those using SGA demonstrates the point.
\end{enumerate}
Moreover, by perusing the results in Table \ref{expres} and the performance maps, one can observe that even  with relatively low computational costs, it is already possible to find high performing parameter settings when an effective meta-optimizer, such as SGA, is applied to explore the learner's parameter space. 


\subsection{High performance in learning contexts}
Table \ref{highperf} introduces an additional measure to assist in comparing learners across learning contexts: the high performance values $HP(k)$. 
As introduced in Section \ref{hpsection}, $HP(k)$ measures how frequent are high performing parameter settings 
within a $(k*100)$\% distance from the maximum on a given performance map. 
HP(k) values thus allow to express in short one of the main insights offered by a performance map: how easy or difficult is to find high performing parameter settings for the learning context.

We could then compare two learning contexts in terms of their  $HP(k)$ values for a given selection of $k$ distances. 
And we could define a learning context as \emph{higher performant} than another if it has higher $HP(k)$ values for a given selection of $k$ distances.

From Table \ref{highperf}, one can observe that the learning contexts with SGA as the meta-optimizer have higher $HP(k)$ values than those associated with Grid Search.
This means that the performance maps associated to SGA contains more parameter settings performing closer to the maximums than performance maps associated with Grid Search.

This finding is due to the capability of genetic algorithms to focus their search towards high performant parameter settings and to avoid low performing ones. On the contrary Grid Search will have to include all parameter settings in its exploration of the parameter space.

In addition, considering the Congressional Voting Records data set, one can note that the learning context with SVM and SGA dominates the learning context with SVM and Grid Search. 
Indeed, Table \ref{highperf} shows that SVM is generally a more robust learner than DT across the considered learning contexts finding consistently higher HP valued performance maps except in the case of Diabetes (the No Free Lunch theorem at works!).

\textbf{Methodology result - In conclusion, the better performing pair <learner, meta optimizer> appears to be <SVM, SGA> over the considered learning contexts.} 

We complete our experimental study by repeating that using a classic performance measures (accuracy, error rate, etc.) together with performance maps and HP values allows for a multi-faceted comparison of learning algorithms across data sets including robustness to varying parameter settings for the learner.

We believe that having more insight on the behavior of a learner is especially useful 
when dealing with novel, unseen data.
Indeed, being able to calculate and possibly visualize its performance map provides more confidence in how the learner would behave in the future and what subset of parameter settings are likely to produce 
high performing outcomes: the highest the HP(k) values, the highest the probability that the learner will operate within the [bestperformance *(1-k), bestperformance] range when variation to its configurations settings will happen in the future.

\begin{table*}[htb]
\caption{High Performance values $HP(k)$)in several learning contexts.  }
\label{highperf}
\begin{tabular}{cccccc}
Data set   & Learner and  & Best Accuracy & HP(0.05)  & HP(0.10)  &  HP(0.20)    \\
  &  Meta Optimization    &               & (within 5\% of best)  & (within 10\% of best)  &  (within 20\% of best)   \\
\hline
Mushrooms & DT - Grid  & 1.00 &   0.16    &  0.16 &	0.66  \\
Mushrooms & DT - SGA    & 1.00 &   0.25    & 0.25 &	0.65   \\
Mushrooms & SVM - Grid & 1.00 &   0.89    & 0.97  &  0.98   \\
Mushrooms & SVM - SGA & 1.00 &   0.89    & 0.93  &  1.00   \\
\hline
Congr. Voting rec. & DT - Grid  & 0.96 &  0.66    &  0.66 &	0.66  \\
Congr. Voting rec. & DT - SGA    & 0.96 &   0.78    & 0.78 &	0.78   \\
Congr. Voting rec. & SVM - Grid & 0.97 &   0.91    & 0.96  &   0.96   \\
Congr. Voting rec. & SVM - SGA & 0.96 &   0.91   & 0.96  &	0.96   \\
\hline
Diabetes & DT - Grid  & 0.75 &   0.12    &  0.15 &	1.00  \\
Diabetes & DT - SGA    & 0.75 &   0.32    & 0.39 &	1.00   \\
Diabetes & SVM - Grid & 0.77 &   0.31    & 0.32  &   0.58   \\
Diabetes & SVM - SGA & 0.77 &   0.30   & 0.30  &	0.57   \\
\hline
Abalone & DT - Grid  & 0.49 &   0.09    &  0.23 &	0.28  \\
Abalone & DT - SGA    & 0.49 &   0.25    & 0.40 &	0.45   \\
Abalone & SVM - Grid & 0.56 &   0.14    & 0.32  &   0.54   \\
Abalone & SVM - SGA & 0.56 &   0.17   & 0.36  &	0.59    \\
\hline
\end{tabular}
\end{table*}		

\section{Conclusions}
In the paper, we propose to map learning algorithms on data (performance map) in order to gain more insights in the distribution of their performances across their parameter space. 
This approach provides useful information when selecting the best configuration for a learning context 
and when comparing alternative learners.
To formalize the above ideas, we introduced the notions of learning context, performance map, and high performance function.
We then applied the concepts to a variety of learning contexts to show their capabilities.

We showed that the proposed methodology can provide more information on the robustness of a learner in a given learning context thus enriching the traditional single-valued performance measures used in literature when comparing learners.

Future research directions are plentiful.
Because meta-optimization is a separate learning task itself, it open up a series of interesting research questions like: how to better use relatively small data samples or data streams.

Another direction is to study the application of this methodology to more sophisticated learning systems such as agent based systems for modeling complex time series in financial applications \cite{NeriExSyst2020,NeriAIcom12,Neri2018755,Neri2019Wivace, neri2021domain, neri2020identify}. 
Or what will happen when neural networks are used as learners? How to select their most important parameters and how to deal with their long training time maybe in control applications \cite{Neri2019Aciids}?

\begin{appendix}

\section{Simple Genetic Algorithm and Grid Search \label{appSGAgs}}
In the paper, we use two meta-optimizer SGA and Grid search.
The pseudo code for the SGA used in this study can be found in Table \ref{SGA} and that of
 Grid search can be found in Table \ref{GS}.

One of the meta-optimization methods used in our work is a Simple Genetic Algorithm (SGA) with elitism \cite{goldberg:book}.
SGA is a well known algorithm therefore we will not explain it in details.
\begin{table}[thb]
\caption{Simple Genetic Algorithm}
\label{SGA} 
{\em \small
\begin{tabbing}
a\=a\=aa\=aa\=aa \kill
\>//Note: each individual codifies for a parameter set for the Learner \\
\>//Function DoExperiment  performs a 10 fold cross validation\\
\>//on Learner, configured with the parameters codified by an individual,\\
\>//applied on the data set Data\\
\\
\>EvaluateFitness(Population, Learner, Data) \\
\>\>for each individual in Population \\
\>\>\>Fitness(individual )= DoExperiment(Learner, individual, Data)\\
\\
\>SGA(PopulationSize, MaxGenerations, Learner, Data) \\
\>\> nGen = 1\\
\>\> BestIndividual = \{\}\\
\>\> Population = initPopulation(PopulationSize, Learner)\\
\>\> EvaluateFitness(Population, Learner, Data)\\
\>\> while nGen < MaxGenerations\\
\>\>\> MatingPool = Select(Population)\\
\>\>\> DoCrossOver(MatingPool)\\
\>\>\> DoMutation(MatingPool)\\
\>\>\> NextGenPopulation = ReplaceIn(MatingPool, Population)\\
\>\>\> EvaluateFitness(NextGenPopulation, Learner, Data)\\
\>\>\> Population = NextGenPopulation\\
\>\>\> Maintain(BestIndividual,Population) //elitism\\
\>\>\> BestIndividual = FindBestSolution(Population)\\
\>\>\> nGen = nGen + 1\\
\>\> end while \\
\>   return(BestIndividual)\\
\end{tabbing}}
\end{table}
\begin{table}[h]
\caption{The Grid Search Algoritm.}
\label{GS} 
{\em \small
\begin{tabbing}
aa\=aa\=aa\=aa\=aa \kill \\
\>//ParameterSpace contains all combinations of parameters for the Learner \\
\\
\>GridSearch(Learner, Data, ParameterSpace) \\
\>\> BestParameterSettings = \{\}\\
\>\> BestAccuracy = 0\\
\>\> for each p in ParameterSpace \\
\>\>\> Accuracy = DoExperiment(Learner, p, Data)\\
\>\>\> if (Accuracy > BestAccuracy) then \\
\>\>\>\> BestParameterSettings = p \\
\>\>   return(BestParameterSettings)\\
\end{tabbing}
}
\end{table}
We implemented SGA in Python 3.8, by adapting the library Genetic Algoritm\footnote{https://pypi.org/project/geneticalgorithm/}.
In particular, we improved the SGA in the library by (1) adding a cache memory inside the fitness function to avoid repeated evaluations of the same individual, and (2) by adding a stopping criterion based on a minimum level of performance. The SGA stops when its best individual has a fitness equal or above the given minimum. 
We did not add these two improvement in the code in Table \ref{SGA} to improve its readability.

The parameters used to run the SGA in all the learning contexts are: max generation = 50, population size = 50, mutation rate = 0.1, crossover rate = 0.9, replacement rate = 0.9, crossover-type = uniform,
stop-when-fitness-is-above = 0.99. 

The second meta-optimization methods used in this work is Grid Search. 
Grid Search consists in enumerating all the possible values inside a given search space and 
in evaluating them.
Also Grid Search is a well known algorithm so we will not comment it.
\end{appendix}

\bibliographystyle{ios1}           
\bibliography{biblio}

\end{document}